\documentclass[10pt,letterpaper]{article}

\usepackage[T1]{fontenc}
\usepackage[utf8]{inputenc}
\usepackage{newtxtext,newtxmath}
\usepackage[margin=0.9in,includefoot]{geometry}
\usepackage{microtype}
\usepackage{mathtools}
\usepackage{setspace}
\usepackage{titlesec}
\usepackage[hang,flushmargin]{footmisc}
\usepackage{hanging}
\usepackage{xurl}
\usepackage[hidelinks]{hyperref}
\usepackage{bookmark}
\usepackage{enumitem}
\usepackage{csquotes}

\hypersetup{
  pdftitle={Language Has Two Parameters: Narrative-Induced Semantic Plasticity and Phase-Sensitive Interpretation},
  pdfauthor={Hollis Robbins},
  pdfsubject={Version 2 manuscript, August 19, 2026},
  pdfkeywords={attention, narrative, linguistic form, semantic plasticity, quantum cognition, phase, monosemanticity, language models, agent memory, reception history}
}
\MakeOuterQuote{"}

\newcommand{\ket}[1]{\ensuremath{\lvert #1\rangle}}

\allowdisplaybreaks[1]
\titleformat{\section}{\large\bfseries}{ }{0pt}{}
\titleformat{\subsection}{\normalsize\bfseries}{ }{0pt}{}
\titlespacing*{\section}{0pt}{1.25\baselineskip}{0.35\baselineskip}
\titlespacing*{\subsection}{0pt}{0.95\baselineskip}{0.25\baselineskip}

\renewenvironment{quote}
  {\list{}{\leftmargin=1.35em\rightmargin=1.35em\topsep=0.3\baselineskip\parsep=0.28\baselineskip\itemsep=0pt}\item\relax}
  {\endlist}

\makeatletter
\renewcommand\footnotesize{\@setfontsize\footnotesize{8.2}{9.4}}
\makeatother

\begin{document}
\thispagestyle{plain}

\begin{center}
  {\fontsize{17}{19}\selectfont\bfseries Language Has Two Parameters:\par}
  \vspace{0.08em}
  {\fontsize{15}{17}\selectfont\bfseries Narrative-Induced Semantic Plasticity and Phase-Sensitive Interpretation\par}
  \vspace{0.48em}
  {\normalsize Hollis Robbins, University of Utah\par}
  \href{mailto:Hollis.Robbins@utah.edu}{Hollis.Robbins@utah.edu}\par
  \vspace{0.18em}
  Version 2 August 19, 2026
\end{center}

\vspace{0.42em}
\noindent\begin{minipage}{\textwidth}
\small
\setstretch{0.96}
\textbf{Abstract.} Reading fiction or encountering narrative generally does not merely add information. The encounter changes the reader. This paper proposes that encounters alter persistent relations among simultaneously active meanings, producing individual and shared histories that population-trained language models do not necessarily retain. A model may be told of an encounter and reproduce its consequences while the history remains in context; this is not the same as being changed by the encounter. This paper formalizes this missing relational state as phase, derives testable predictions about encounter order, quotation, and suppressed meanings, and argues that future AI agents will need persistent semantic states indexed to particular individuals and relationships. The matching risk is semantic poisoning: an attack that re-signs relations among meanings already present.
\end{minipage}

\vspace{0.35em}
\noindent\textbf{Keywords:} attention; narrative; linguistic form; semantic plasticity; quantum cognition; phase; monosemanticity; language models; agent memory; reception history

\vspace{0.25em}
\section{1. The problem}\label{the-problem}

Long before the transformer era, poetry was described as the careful configuration of words, each of which has a constellation of associations.\footnote{I. A. Richards, \emph{The Philosophy of Rhetoric} (New York: Oxford University Press, 1936) claims words are laden with meanings from past contexts and animate each other in combination. William Empson, \emph{Seven Types of Ambiguity} (London: Chatto and Windus, 1930), also claims that multiple meanings are live simultaneously; the poetic function of forms is extended in Jakobson (note 6). W. S. Graham, ``Notes on a Poetry of Release,'' \emph{Poetry Scotland} 3 (July 1946), states the configuration claim and the reader claim together: a poem is made of words, words of a certain order, good or bad by the significance of its addition to life; it is brought to life by the reader and takes part in the reader's change. The transformer era refers to Ashish Vaswani et al., ``Attention Is All You Need,'' \emph{Advances in Neural Information Processing Systems} 30 (2017): 5998--6008. The earlier distributional tradition includes Zellig S. Harris, ``Distributional Structure,'' \emph{Word} 10 (1954): 146--162; J. R. Firth, ``A Synopsis of Linguistic Theory 1930--1955,'' in \emph{Studies in Linguistic Analysis} (Oxford: Blackwell, 1957); Tomas Mikolov et al., ``Distributed Representations of Words and Phrases and Their Compositionality,'' \emph{Advances in Neural Information Processing Systems} 26 (2013).} Line lengths, rhyme, and form also carry constellations of associations. The coexistence of these constellations causes amplitude and phase shifts in meaning. Every word carries a cloud of associations in orbit around it, which poets refer to as a constellation: meanings that exert a changing gravitational pull on other words depending on words nearby. The difficulty many people have with poetry is that meaning is hard to discern and poetry can mean different things to different people. Words, vectors, and cosine similarity are the ordinary machinery. All of it estimates the first parameter. This paper is about the second, and what it would take for a language model to represent it. The two parameters are different kinds of quantity; neither can be computed from the other. The first parameter, amplitude, is about the strength and value of meaning by itself. The second parameter, phase, is about the relations of meanings to other meanings: how two or more meanings can be live simultaneously, can combine, can reinforce or cancel one another. Phase can change while the strengths of meaning stay the same; a meaning can be present at full strength with its contribution reversed.

You can think of all fiction generally---including poetry, novels, film, drama---as featuring not only language but also characters, genres, plots, histories, retrospective reinterpretations, countless other features. Fictional characters develop their own constellations and become embedded in individual and cultural consciousnesses, gathering their own constellations of associations and entering the constellations of subsequent fiction and poetry. Genres and plots have this property. Encountering fiction reorganizes the representational capacities of readers and it affects the statistical properties of literary language.

As an example: midway through \emph{The Terminator} (1984), James Cameron's film about a cyborg assassin (played by Arnold Schwarzenegger) sent back from the future to kill a specific target, the cyborg sits in a rented room repairing its wounds.\footnote{\emph{The Terminator}, directed by James Cameron (Orion Pictures, 1984).} The decaying organic matter smells. A janitor knocks, asking if there is a dead cat in there. The film cuts to the cyborg's point of view. A text menu labeled POSSIBLE RESPONSE shows a list of six options; a cursor moves down and stops. The cyborg says, "FUCK YOU, ASSHOLE." It's a laugh line.

To most speakers of American English, the phrase is hostile. But individuals who have seen the film can use the same phrase, perhaps spoken in a flat, Austrian-tinged cadence, to express recognition and affection. They also know that those who have not seen the film might overhear and assume hostility. A robust theory of meaning must explain how a meaning can be simultaneously suppressed, preserved, and mean its opposite. Think of the phrase as the minimal case of linguistic form and a poem as the maximal case.

Today's viewer is likely familiar with autocomplete, if not chatbots. The cyborg's menu is now recognizable.\footnote{The historical variability of interpretation has a long critical genealogy. See Hans Robert Jauss, ``Literary History as a Challenge to Literary Theory,'' trans. Timothy Bahti, \emph{New Literary History} 2, no. 1 (1970): 7--37; Stanley Fish, \emph{Is There a Text in This Class? The Authority of Interpretive Communities} (Cambridge, MA: Harvard University Press, 1980). My theory adds a formal claim about semantic states linked to history.} The phrase now can be used between two people familiar with the film, in the context of AI, to mean ``you are using an old model.'' The development of a new semantic system is a problem that current theories of machine intelligence do not account for. Interpretive architecture is not fixed. People understand meaning differently over time. Narrative-comprehension studies show sharing an interpretation can produce similar behavioral and high-level neural responses and that changes in functional connectivity can follow.\footnote{Yaara Yeshurun et al., ``Same Story, Different Story: The Neural Representation of Interpretive Frameworks,'' \emph{Psychological Science} 28 (2017): 307--319; Mai Nguyen, Tamara Vanderwal, and Uri Hasson, ``Shared Understanding of Narratives Is Correlated with Shared Neural Responses,'' \emph{NeuroImage} 184 (2019): 161--170; Gregory S. Berns et al., ``Short- and Long-Term Effects of a Novel on Connectivity in the Brain,'' \emph{Brain Connectivity} 3, no. 6 (2013): 590--600.}

This paper claims that no amount of additional context-free corpus pretraining---language statistics from large bodies of text alone, without records of who encountered texts when, in what order, and with whom---will model the phenomenon of a phrase meaning one thing to most speakers, another thing between particular speakers, and something else again to one or both over time. This kind of pretraining would continue to miss the history of transformations undergone by an individual semantic system and an individual's model of others' histories. Both are largely absent from every corpus; individual accounts of this kind are rarely written down at scale.\footnote{Herbert H. Clark, \emph{Using Language} (Cambridge: Cambridge University Press, 1996). Clark is helpful on joint perceptual experience and common ground.} In Section 3.4, I prove that a population average never determines what a word or phrase means to one person or between two, at any scale, and I state as a premise that what averaging destroys is the arrangement. The later sections extend the argument to audiences and to language models themselves.

\section{2. Propositions}\label{propositions}

The theory has seven propositions.

\begin{quote}
\textbf{Proposition 0 (Capacity).} Language allows a speaker to set phase in the signal. This means that an utterance can transmit a relation between live meanings through cadence, meter, rhyme, lineation, quotation, or genre.\footnote{Roman Jakobson, ``Closing Statement: Linguistics and Poetics,'' in \emph{Style in Language}, ed.~Thomas A. Sebeok (Cambridge, MA: MIT Press, 1960), 350--377, describes the poetic function as the projection of equivalence from the axis of selection into the axis of combination. Charles F. Hockett, ``The Origin of Speech,'' \emph{Scientific American} 203 (1960): 88--96. Herbert H. Clark and Richard J. Gerrig, ``Quotations as Demonstrations,'' \emph{Language} 66 (1990): 764--805.} Language allows it, fiction installs it, history distributes it.

\textbf{Proposition 1 (Plasticity).} Fiction changes the semantic system of the reader. New constellations form, old meanings gain or lose weight, the relations among them are reset.

\textbf{Proposition 2 (Two parameters).} These changes fix two properties of the constellation that later language activates. Amplitude is how strongly a meaning contributes to interpretation. Phase is how coactivated meanings combine, whether they reinforce or cancel. Both properties belong to an individual's state and persist between encounters. Neither belongs to the word itself.

\textbf{Proposition 3 (History dependence).} Interpretation is a function of an individual's encounter history (what fictional works were read in what order) and of that person's model of other people's encounter histories. Formally: individuals with different histories interpret the same words differently.

\textbf{Proposition 4 (Order dependence).} The order of encounters changes the resulting state. Two persons with the same encounters in different orders will have different constellations of associations. A later encounter recomputes the stored relations of an earlier one; this is retrospective reinterpretation. Formally: semantic updates do not commute.

\textbf{Proposition 5 (Coordination).} A shared encounter creates partially coordinated constellations among communities. The coordination allows allusion, quotation, irony, and cultural recognition. Individuals can share a constellation of associations without holding any belief in common.

\textbf{Proposition 6 (Preservation under suppression).} Suppression can preserve meaning. When a speaker suppresses the dominant meaning of a phrase (the \emph{Terminator} example), the suppressed component remains active and combines with the expressed components. What changes is its phase relation. Marked ironic quotation is an extreme case where the suppressed meaning is present at full strength and its contribution is reversed.

\textbf{Corollary (Historical situatedness and reflexivity).} An audience is a distribution of encounter histories, and that distribution changes over time. Later technologies and narratives can rephase earlier works. Propositions 1 and 3 through 5 therefore apply to the production of theories as well as to the reception of utterances.
\end{quote}

\section{3. The formal theory}\label{the-formal-theory}

Ordinary probability cannot represent two things the propositions require. A meaning that a speaker suppresses must still affect an interpretation, with its relation to the other live meanings reversed. And the order of an individual's encounters must be allowed to change the result. Quantum probability supplies both a phase relation whose effect appears only when two meanings are live at once and operations whose order changes the state. I am drawing on quantum probability for those two properties; I'm claiming nothing about physics, only about representational structure.\footnote{Jerome R. Busemeyer and Peter D. Bruza, \emph{Quantum Models of Cognition and Decision} (Cambridge: Cambridge University Press, 2012). My paper does not claim that the brain is a quantum computer, that linguistic meaning is physically quantum, or that complex numbers are the only possible implementation. I claim only that a formalism separating semantic strength from the relation among coactive meanings captures phenomena that amplitude-only descriptions obscure.} A rich enough classical system could reproduce the behavior with signed variables and order-sensitive transitions, and it would still have to represent the second parameter and the history-indexed transformations of a constellation at issue here.

\subsection{3.1 States}\label{states}

Represent an individual's semantic system for a given phrase \emph{w} as a state in a meaning space. Write the distinguishable meanings the system can produce as \ket{m_1}, \ket{m_2}, \ket{m_3}, using Dirac's notation, in which \ket{m} labels a basic meaning state. For the phrase above: \ket{m_1} = insult, \ket{m_2} = quotation of \emph{The Terminator}, \ket{m_3} = generic defiance. The state of an individual with encounter history \emph{h} is a weighted combination:

\[\ket{\psi_h}=\sum_i a_i e^{i\theta_i}\ket{m_i},\qquad \sum_i a_i^2=1.\]

This is the constellation of associations. The \ket{m_i} are what it holds, \(a_i\) is how bright each one is, and \(\theta_i\) is where it stands relative to the rest.

Each live meaning \emph{i} is specified by two parameters: an amplitude \(a_i\) and a phase \(\theta_i\). The amplitude measures how strongly that meaning is present. The phase measures where it stands relative to the other live meanings, which fixes how it combines with them when several are active at once. A constellation contains one such pair for each live meaning, and only phase differences affect interpretation. Classical accounts of meaning work with connection weights, reading probabilities, and association strengths, and each of these supplies the first parameter. A signed connection can lower a competitor's amplitude; none reverses what a meaning contributes while that meaning stays fully present. Proposition 2 asserts the second, and everything distinctive in the theory follows from it.

The combination above is a superposition, which differs from a mixture. A mixture says the individual holds the insult meaning with probability \(a_1^2\) or the quotation meaning with probability \(a_2^2\), and we do not know which. A superposition says both meanings are live at once, with a definite phase relation between them. The two descriptions make different predictions. Section 5 gives the experiments that would tell them apart.

When you hear a phrase, several meanings light up in your head at once. Which ones light up depends on what you have read, watched, and heard. Someone who has seen a particular show or read a particular book has a meaning available that someone who hasn\textquoteright{}t does not. New inputs can put an old meaning in play or add one that nobody had before. As Jorge Luis Borges has noted, the order in which we encounter books can seem to reverse the direction of literary influence.\footnote{Jorge Luis Borges, ``Kafka and His Precursors,'' trans. Eliot Weinberger, in~\emph{Selected Non-Fictions}, ed. Eliot Weinberger (New York: Viking, 1999), 365.} An encounter can bring two meanings into relation that had nothing to do with each other. So the meanings a word or phrase can set off, and how they relate to one another, belong to the person and to what that person has read or encountered. Each person has a constellation that may have nothing to do with a dictionary.

The expression has a state as well. Write \ket{\psi_w} for the phrase as configured: the constellations of its words and the devices of its form. The devices constrain the relative phases an utterance transmits; quotation marking, cadence, meter, rhyme, and lineation fix relations among the couplings defined in Section 3.2. Interpretation is an interaction between two states, the phrase as configured and the person as their history has left them. Agostino and colleagues index the state to the expression and treat the agent as the measurement; the present theory requires both indices (Proposition 0).

\subsection{3.2 Interpretation as measurement}\label{interpretation-as-measurement}

Interpreting is measuring. You put a question to the constellation---What do these words mean? What is this speaker doing?---and the arrangement gives an answer with some probability. Let O be the outcome you are asking about, and write

\[c_i=k_i e^{i\phi_i}=\langle O\mid m_i\rangle\]

for the contribution of meaning \(m_i\) to that outcome under the phrase's configured form, where \(k_i\) is its magnitude and \(\phi_i\) its phase. With two meanings live,

\[P(O)=\left|a_1k_1+a_2k_2e^{i\Delta\theta}\right|^2,\]

where \(\Delta\theta = \left( \theta_{2} - \theta_{1} \right) + \left( \phi_{2} - \phi_{1} \right)\). The first difference comes from the person's history. The second is set by the form of the signal---the cadence, the quotation marking, the lineation---which is Proposition 0 in the equation. A single judgment gives only their sum. Prediction 4 varies the history and holds the signal fixed; Prediction 6 varies the signal and holds the history fixed. When the two meanings contribute equally, \(k_{1} = k_{2} = 1/\sqrt{2}\), and

\[P(O)=\frac12\left(a_1^2+a_2^2+2a_1a_2\cos\Delta\theta\right).\]

The first two terms are what each meaning contributes on its own. The third is interference, and it exists only because both are live at once. A mixture gives you the first two and stops.

That third term is Proposition 6 written as mathematics. Holding a meaning back leaves \(a_2\) alone and moves \(\Delta\theta\). Set \(\Delta\theta=\pi\), and \(\cos(\Delta\theta)=-1\), so the cross term is \(-2a_1a_2\): the held-back meaning contributes at full strength with its sign reversed. That is the affectionate greeting between two people who have both seen the film.

The same term, the interference term, might be seen as the mechanism for what W. E. B. Du Bois describes as double-consciousness, the sense of looking at oneself through the eyes of others, for what Mikhail Bakhtin calls double-voiced discourse, two voices audible in one utterance, for the double-voiced utterance that Henry Louis Gates Jr. theorizes as Signifyin(g), and for what relevance theory calls the echoic character of irony while leaving the echo unanalyzed.\footnote{Mikhail Bakhtin, ``Discourse in Dostoevsky,'' in \emph{Problems of Dostoevsky's Poetics}, ed.~and trans. Caryl Emerson (Minneapolis: University of Minnesota Press, 1984); W. E. B. Du Bois, \emph{The Souls of Black Folk} (Chicago: A. C. McClurg, 1903), ch. 1. Henry Louis Gates Jr., \emph{The Signifying Monkey: A Theory of African-American Literary Criticism} (New York: Oxford University Press, 1988), formalizes the double-voiced utterance in African American expression as Signifyin(g). Dan Sperber and Deirdre Wilson, ``Irony and the Use-Mention Distinction,'' in \emph{Radical Pragmatics}, ed.~Peter Cole (New York: Academic Press, 1981); Deirdre Wilson and Dan Sperber, ``Explaining Irony,'' in \emph{Meaning and Relevance} (Cambridge: Cambridge University Press, 2012).} Free indirect discourse of the kind Jane Austen pioneered is the most refined version of the device. The thought is the character's and the grammar is the narrator's, and nothing marks the switch---no quotation marks, no "she thought." Both are live simultaneously; grammar alone tells how they are related. Critics have argued for a century over whether such sentences carry two voices, one, or none. The argument is a choice among superposition, mixture, and collapse, made without the second parameter---the relation between two meanings that are live at once. Semantics has its own two-parameter treatment of these sentences: a context of utterance and a context of thought, developed by Schlenker and extended by Sharvit and Eckardt. A sentence is evaluated from those two points. Phase---a relation between two meanings live in one reader---belongs to a person rather than to a sentence. A second parameter is needed to reverse what a meaning contributes while the first meaning is still fully present.\footnote{Charles Bally, ``Le style indirect libre en fran\c{c}ais moderne I,'' \emph{Germanisch-Romanische Monatsschrift} 4, no. 10 (1912): 549--556; and ``Le style indirect libre en fran\c{c}ais moderne II,'' \emph{Germanisch-Romanische Monatsschrift} 4, no. 11 (1912): 597--606; Roy Pascal, \emph{The Dual Voice: Free Indirect Speech and Its Functioning in the Nineteenth-Century European Novel} (Manchester: Manchester University Press, 1977); Ann Banfield, \emph{Unspeakable Sentences: Narration and Representation in the Language of Fiction} (Boston: Routledge \& Kegan Paul, 1982); Dorrit Cohn, \emph{Transparent Minds: Narrative Modes for Presenting Consciousness in Fiction} (Princeton: Princeton University Press, 1978). Joe Bray, ``The `Dual Voice' of Free Indirect Discourse: A Reading Experiment,'' \emph{Language and Literature} 16, no. 1 (2007): 37--52, finds readers divide in attributing such sentences. A conversation with Peter Danenberg in March 2026 encouraged me to think of FID in engineering terms.}

The architecture now standard in language modeling does not give the interference term explicit representation. In the transformer, the coefficients that relate one position to another are the outputs of an averaging mechanism: the coefficients themselves are nonnegative and designate no inversion, while signed value vectors and projections downstream can realize cancellation: \(\operatorname{Attention}(Q,K,V)=\operatorname{softmax}\!\left(QK^{\mathsf T}/\sqrt{d_k}\right)V\), a weighted sum in which every weight is a nonnegative real number and the weights sum to one.\footnote{Ashish Vaswani et al., ``Attention Is All You Need,'' \emph{Advances in Neural Information Processing Systems} 30 (2017): 5998--6008. The attention output is defined as a ``weighted sum of the values'' (their section 3.2), with weights produced by a softmax; the positional encodings are defined in section 3.5.} The genealogy of this signlessness: attention descends from translation alignment, where a source token corresponds to a target token to some degree and anti-correspondence has no meaning, and parallel text presents two codes live at once with their relation pinned at equivalence. The relation between the two codes is fixed at equivalence before any model sees the data, and the operation survived the founding task.\footnote{Dzmitry Bahdanau, Kyunghyun Cho, and Yoshua Bengio, ``Neural Machine Translation by Jointly Learning to Align and Translate,'' Proceedings of the 3rd International Conference on Learning Representations (2015). Vaswani et al. retained the alignment operation while removing the recurrence around it.} The network is real-valued end to end, and a complex amplitude has an exact representation in two real coordinates, so the deficit is not expressive impossibility: no parameter is designated as the relation, none persists past the sequence, and none is indexed to an individual agent. The cross term \(2a_{1}a_{2}\cos(\Delta\theta)\) therefore has no dedicated representation, and whatever sign structure a trained model exhibits is distributed across shared weights---which returns the problem to Section 3.4, since shared weights are population objects with no agent index.

The reversed sign depends on what you measure. Measure what the words mean, and every meaning in the constellation is still there at full strength. Measure what the speaker is doing to the person listening, and the same constellation, arranged as it is, gives a different answer. Nothing in the constellation has changed between the two measurements. The arrangement is what produces the difference.

\subsection{3.3 Encounters and their order}\label{encounters-and-their-order}

An encounter---reading a novel, seeing a film, hearing a phrase quoted at a party---changes the constellations a person carries, and a life is one encounter after another in a particular order. Write the change an encounter \emph{F} makes as a map \(\Phi_F\), and a history as a composition of those maps:

\[\rho_h=\Phi_{F_n}\circ\cdots\circ\Phi_{F_1}(\rho_0).\]

The state here is written \(\rho\) rather than \ket{\psi} because people forget, and because anyone modeling another person works with partial knowledge; \(\rho\) is the standard way to represent a state that is mixed or only partly known.

Proposition 4 says these maps do not commute:

\[\Phi_{\mathrm{source}}\circ\Phi_{\mathrm{use}}\neq\Phi_{\mathrm{use}}\circ\Phi_{\mathrm{source}}.\]

Someone who has heard a phrase used by someone for years and only later encounters the work it comes from may reinterpret everything they heard before. Someone who encounters the work first was never in that position. Both have the same encounters and different constellations. The same holds when the later encounter is a life event or technology rather than a work: what was ordinary or strange when a work first appeared can be strange or ordinary by the time a later audience arrives, and that audience brings a constellation in which the strangeness has already been placed.\footnote{I am thinking here about the opening scene of the Coen Brothers' film \emph{The Big Lebowski} (1998) where Jeffrey ``the Dude'' Lebowski dates a check September 11, which post-2001 viewers often find deeply meaningful.}

In a simple Bayesian model, each piece of evidence is interchangeable, so the order of arrival does not affect the result. More general Bayesian and predictive-processing models can let earlier experiences change how later ones are processed, and they can represent path dependence, plasticity, and order effects.\footnote{Karin Kukkonen, \emph{Probability Designs: Literature and Predictive Processing} (New York: Oxford University Press, 2020).} My claim is narrower than a rejection of those frameworks. Any adequate account has to represent noncommuting transformations of semantic architecture across a life. Update semantics and dynamic semantics already model noncommutative change at the scale of a conversation, where what matters is the order of sentences.\footnote{Frank Veltman, ``Defaults in Update Semantics,'' \emph{Journal of Philosophical Logic} 25 (1996): 221--261; Jeroen Groenendijk and Martin Stokhof, ``Dynamic Predicate Logic,'' \emph{Linguistics and Philosophy} 14 (1991): 39--100.} In quantum probability, order-dependence is built into the operations rather than added to them, and the calculus constrains how state, order, and measurement relate. The biographical case remains largely unformalized.

\subsection{3.4 What pretraining cannot recover}\label{what-pretraining-cannot-recover}

Grant the model an exact estimate of the text-only law its corpus supports. That law is an average across everyone:

\[\bar{\rho}=\sum_h p(h)\rho_h,\]

the sum over encounter histories weighted by their population frequency. The average is taken in a common space that includes every person's constellation, so a meaning available to some people and not others shows up as strength on something the others lack. Strength sits on the diagonal of \(\rho\), one entry per meaning. Arrangement sits off the diagonal, one entry per pair. For two meanings, write the off-diagonal entry for history \(h\) as \(\left( \rho_{h} \right)_{12} = r_{h}e^{- i\Delta\theta_{h}}\), with the conjugate in \(\left( \rho_{h} \right)_{21}\). In the pure-state case, \(r_{h} = a_{1}a_{2}\); in a mixed state, \(r_{h}\) is the magnitude of the off-diagonal relation that remains. Its average across the population is

\[\bar{\rho}_{12}=\mathbb{E}\!\left[r_h e^{-i\Delta\theta_h}\right].\]

Call a corpus agent-deindexed when it retains what was written and not who encountered it, in what order, or with whom---which is every corpus now in use. Different readers, different orders, different lives: the arrangements point every which way, and conditional on what an agent-deindexed corpus retains they cancel, \(\mathbb{E}[r_h e^{-i\Delta\theta_h}\mid X]\approx0\). Averaging deletes the arrangement and keeps the meanings. Call the cancellation the dephasing premise. It is an empirical claim about populations, and the proposition below does not depend on it.

\begin{quote}
\textbf{Proposition (non-identifiability of histories from the marginal).} Assume (i) every history-specific state \(\rho_h\) is embedded in the common ambient basis of Section 3.1; (ii) the measurement \(E_O(w)\) is fixed given the variables \(X\) an agent-deindexed corpus retains. Then the corpus-supported law determines at most \(\bar{\rho}_X=\mathbb{E}[\rho_h\mid X]\), with \(P(O\mid w,X)=\operatorname{Tr}[E_O(w)\bar{\rho}_X]\) by linearity of the trace. A density matrix does not determine the ensemble that realizes it: infinitely many assignments \(\{p(h),\rho_h\}\) produce the same \(\bar{\rho}_X\).\footnote{Michael A. Nielsen and Isaac L. Chuang, \emph{Quantum Computation and Quantum Information}, 10th anniversary ed. (Cambridge: Cambridge University Press, 2010), Theorem 2.6, state the pure-state case as unitary freedom in the ensemble; decompositions into mixed states inherit the non-uniqueness.} However exactly \(\bar{\rho}_X\) is estimated, nothing in it assigns a history-conditioned state to a particular agent or dyad. Under the dephasing premise, \(\bar{\rho}_X\) is in addition approximately diagonal in that basis: meanings with population frequencies and no arrangement.
\end{quote}

A larger corpus sharpens the estimate of \(\bar{\rho}\); the proposition says no amount of sharpening will recover any particular constellation.\footnote{Francesco Corielli, "When Is Next-Token Prediction Useful? Marginalization, Ergodicity, Mixture Identifiability, Local Sufficiency, RAG, Tools, and Programming," arXiv:2605.23278 (2026), and "The Need for an External Observer: Formalizing the Sufficiency Gap," arXiv:2605.26711 (2026). Corielli separates the full conditional law \(p_{\mathrm{full}}(x_{t+1}\mid x_{\le t},z_t)\) from the marginal text-only law obtained by integrating out \(z\), and states the usefulness condition as \(I(X_{t+1};Z_t\mid X_{\le t})\approx0\). His \emph{z} is the circumstance of production: world state, facts, speaker beliefs, goals, audience, task constraints. Mine is the encounter history of the receiver. His marginalization discards information about a situation. Mine discards a relation between two meanings held by one person.} This is the pretraining claim: what the corpus gives you does not determine what any particular person or pair holds. The corpus does contain the sources, the quotations, and the commentary, so a quotation reading enters the trained weights at the population level. Weights with no record of who read what can learn general patterns. What they cannot do is say which constellation belongs to these two people, with these histories, in this order, because nothing in the record links them.\footnote{Christopher J. Agostino, Quan Le Thien, Molly Apsel, Denizhan Pak, Elina Lesyk, and Ashabari Majumdar, ``A Quantum Semantic Framework for Natural Language Processing,'' in \emph{Quantum AI and NLP: First International Conference, QNLPAI 2025, Bloomington, IN, USA, August 6--9, 2025, Proceedings}, ed. Damir Cavar, Vaneet Aggarwal, Jerome Busemeyer, and Samuel Yen-Chi Chen, Communications in Computer and Information Science 2659 (Cham: Springer, 2026), 134--155, \url{https://doi.org/10.1007/978-3-032-13883-5_8}.}

A model given a dyad's history in its context window---these two individuals both saw this film, each knows the other did---can often simulate the expected interpretation. The prompt states the missing variables propositionally. It does not establish that the model has undergone a durable transformation, that the effects of the encounter persist after the prompt disappears, or that the model maintains an evolving representation of how a particular individual was changed. Context can cue a simulation of \(\rho_h\); plasticity is something else. Narrative-induced semantic plasticity, the subtitle's term, is the durable rearrangement of a reader's constellation by an encounter with a story: the relations among live meanings change, the strengths need not, and the change outlasts the story's presence. It requires a state that remains altered across later encounters. The distinction separates in-context conditioning from the phenomenon the theory describes.\footnote{Corielli, "When Is Next-Token Prediction Useful?" \S15, argues the stronger point that a prompt selects among learned conditional behaviors and cannot install one that was not learned, so a fully informative prompt fails when the conditioning pattern is absent from the training support. The companion paper states the same requirement as learned conditional availability (arXiv:2605.26711, Assumption 2).}

Imagine a community as a set of people whose constellations were arranged by the same encounters, who read the same books, watched the same shows, sat in the same classrooms, and so hold the same meanings in the same relations. What binds them is their semantic encounters rather than hometown or politics, say. Every person belongs to many such groups at once.

Define the community-conditioned state

\[\rho_C=\mathbb{E}[\rho_h\mid C],\]

the average over histories within an interpretive community C, and a community coherence

\[\kappa_C=\mathbb{E}[e^{i\Delta\theta}\mid C].\]

The global population can have \(\kappa\approx0\) while a community holds \(|\kappa_C|>0\): phases uncoordinated across the whole population remain coordinated among those who share the relevant encounters, and shared viewing, teaching, quotation, and re-release sustain the coordination. The dephasing premise then shows its limit. Global averaging does not show that phase is absent from individuals; it conceals phase relations that remain coherent within historically organized communities, and a diagonal \(\bar{\rho}\) is compatible both with incoherent individuals and with coherent communities whose phases cancel in aggregate. Reception history acquires a variable: an interpretive community in Fish's sense (note 3) is a set of histories with \(|\kappa_C|>0\) for the relevant texts, and a canon is coherence maintained by institutions.

The continuous production of new teenage slang is helpful here.\footnote{Hollis Robbins, ``AI and the Humanities,'' interview by Amanda Anderson, \emph{Meeting Street: Conversations in the Humanities}, episode 14 (Cogut Institute for the Humanities, Brown University, December 2023), \url{https://humanities.brown.edu/media/meetingst/14}.} A frozen population model has no live slang because there can be no live slang in a population average: new terms work by re-signing a common token, legible to holders of a shared history and read at the marginal by everyone else. The value of the term is the gap between the community's conditional reading and \(\bar{\rho}\): slang is \(|\kappa_C|>0\) engineered against \(\kappa\approx0\). By the time a model catches up with a new term and can deploy it, it is long past, and the community has rephased another token. The model reads at \(\bar{\rho}\) unless the prefix identifies the community, and slang is language engineered to fail at \(\bar{\rho}\).\footnote{Corielli, "When Is Next-Token Prediction Useful?" \S7, gives the condition as mixture identifiability: \(p_{\mathrm{mix}}(x_{t+1}\mid x_{\le t})=\sum_k p(k\mid x_{\le t})p_k(x_{t+1}\mid x_{\le t})\). A model reads at the local regime law when the prefix identifies the regime and at the blended average when it does not. His regimes are genres with textual signatures, such as \texttt{def merge\_sort(arr):} or \texttt{\textbackslash begin\{proof\}}. An interpretive community has no such signature, and slang is constructed so that a prefix legible to members reads as ordinary language to everyone else.} Spending on faster ingestion buys a faster estimate of \(\bar{\rho}\). Slang competence is relational; competence requires the agent-indexed state of Section 6. Slang is the cheapest test of the architectural program. It works by the same intimacy as the quoted insult: a term is live for a community that shares a history and reads as something else to everyone else. The model absorbs the words belatedly, when they are no longer live, and never has the history that made them live.

\subsection{3.5 Decoherence: live allusion, dead idiom, recovery}\label{decoherence-live-allusion-dead-idiom-recovery}

The off-diagonal entries decay when nothing sustains them:

\[\rho_{12}(t)=a_1a_2e^{-i\Delta\theta}e^{-t/\tau}.\]

Relations between two meanings have to be renewed, which generally occurs by rereading, by teaching, by people quoting the work to each other, by a reissue that puts it back in front of an audience. Without renewal relations may fade, and \(\tau\) measures how long that takes. The equation above is the simplest case, where nothing renews it; when something does, the relation strengthens again. What eventually decays is the relation, not the meanings. Both meanings survive at full strength but nothing connects them any longer. The equation is a proposed model rather than an established law. The point is separating the fading of a relation from the loss of a meaning.

Three regimes can be distinguished. A live allusion is a coherent superposition: source and use are phase-related, and the interference is audible. A dead idiom is a mixture: ``hoist with his own petard'' retains its use-meaning, while its source and literal meanings remain available but no longer interfere with that use, even for readers familiar with \emph{Hamlet} and the literal sense (a petard is a small siege bomb, and the line describes an engineer blown up by his own device). Scholarly recovery is re-preparation of coherence. Oh and colleagues traced idiom comprehension in four transformers and found both readings still live at the output, carried on separate routes: an intermediate path favoring the figurative, a direct path favoring the literal. That is coexistence in a model. The two readings can coexist; irony requires the literal reading to work against the figurative.\footnote{Tug of War: Soyoung Oh, Xinting Huang, Mathis Pink, Michael Hahn, and Vera Demberg, "Tug-of-war between idioms' figurative and literal interpretations in LLMs," in \emph{Proceedings of the 19th Conference of the European Chapter of the Association for Computational Linguistics}, vol. 1, \emph{Long Papers} (2026), 2942--2958. Their causal tracing covers 245 idioms in Llama3.2-1B, with replication in Llama3.1-8B, Qwen2.5-0.5B, and Qwen2.5-7B.} The model predicts an intermediate regime of partial coherence, measurable as weakened irony, and it gives cultural transmission a quantitative variable: \(\tau\) for Shakespeare's petard ran out among general speakers some generations ago; \(\tau\) for Schwarzenegger's phrase is maintained by re-release, sequels, and continuous quotation. Dead metaphor is decohered meaning; decohered differs from forgotten.\footnote{I. A. Richards, \emph{The Philosophy of Rhetoric} (New York: Oxford University Press, 1936), objects that ``dead'' misleads because revival is always available; George Lakoff, ``The Death of Dead Metaphor,'' \emph{Metaphor and Symbolic Activity} 2 (1987): 143--147, argues that conventional metaphors remain alive as mappings. The present distinction separates the claims: the components persist as weights, which is what both observations track, and the relation decays, which is what ``dead'' names. Persistence of the components is why recovery succeeds. Aristotle defines metaphor as transfer (\emph{Poetics} 21, 1457b) and observes that ordinary conversation is full of unnoticed metaphors (\emph{Rhetoric} 3.2, 1404b); neither text has an account of the decay.}

Later encounters can also create relations that did not exist before. A work can acquire relations to technologies and concepts that postdate it, which makes reception history a record of how audiences were prepared and prepared again.

\subsection{3.6 The dyad}\label{the-dyad}

Interpretation between individuals requires one further level. Individual A interprets individual B's utterance by conditioning the measurement on A's model of B's state, written \({\widehat{\rho}}_{B}^{(A)}\), and B produced the utterance with a model of A, recursively. This recursion is the Rational Speech Acts model, in which speakers and listeners reason about each other's reasoning.\footnote{Michael C. Frank and Noah D. Goodman, ``Predicting Pragmatic Reasoning in Language Games,'' \emph{Science} 336 (2012): 998.} Hawkins and colleagues extend that lineage to the structure at issue here. In their hierarchical account, a pair builds partner-specific meaning through interaction, and community conventions are priors abstracted from many partners, so an individual, a dyad, and a population are separate levels of one model.\footnote{Robert D. Hawkins, Michael Franke, Michael C. Frank, Adele E. Goldberg, Kenny Smith, Thomas L. Griffiths, and Noah D. Goodman, "From partners to populations: A hierarchical Bayesian account of coordination and convention," \emph{Psychological Review} 130, no. 4 (2023): 977--1016. Old words take new partner-specific meanings by inference from the interaction at hand rather than from a history of encounters with works. Nothing in their classical model holds two meanings live at once with a relation between them.} RSA already contains the modeling of Proposition 3. Standard RSA operates on nonnegative probabilities over utterances, meanings, and utilities. It can increase or decrease the likelihood of a reading and can represent negative utilities; it does not ordinarily represent two live meanings together with the phase relation between them. A classical model that adds a second variable, one that can carry a sign, may reproduce the effect. I claim that the second variable is necessary and I represent it as phase.

Sharing an encounter does not mean holding the same constellation. Two people can share enough to recognize a quotation and still arrange the meanings differently, because one met the work before a later technology existed and the other met it after. They agree on the source and disagree about what makes the quotation funny. This is why Proposition 5 says partial coordination rather than shared meaning.

\section{4. The boundary case outside language: Move 37}\label{the-boundary-case-outside-language-move-37}

Go is a board game in which two players alternately place black and white stones on the intersections of a 19-by-19 grid, each attempting to surround more territory than the opponent. Professional play observes conventions with the standing of grammar: stones on the third line from the board's edge secure territory, stones on the fourth line project influence toward the middle, and a ``shoulder hit''---a stone placed diagonally against an opponent's stone---is a standard reducing technique at those heights. Received professional judgment held a shoulder hit at the fifth line to be bad play.

AlphaGo, a program built by DeepMind, selected moves with two neural networks and a tree search. Its policy network was trained on roughly thirty million board positions from expert human games to predict the move a human would play; a further training stage improved the policy through millions of games of self-play, the program playing against copies of itself.\footnote{David Silver et al., ``Mastering the Game of Go with Deep Neural Networks and Tree Search,'' \emph{Nature} 529 (2016): 484--489.} In March 2016 in Seoul, AlphaGo played a five-game match against Lee Sedol, winner of eighteen international titles, and won four games to one. The thirty-seventh move of the second game was a fifth-line shoulder hit. The probability that a human professional would play the move was estimated at about one in ten thousand. The commentating professionals were audibly confused; Fan Hui, the European champion who had trained against the program, said, ``It's not a human move.''\footnote{The estimate and Fan Hui's remark are documented in \emph{AlphaGo}, directed by Greg Kohs (2017), and in DeepMind's match commentary.}

Move 37 is the theory's boundary case: a semantic system with no words, in which phase exists only in the receivers. This case demonstrates Propositions 1, 3, and 5.

\textbf{First,} an individual trajectory produces meanings absent from the population corpus. The human game corpus rated the move at one in ten thousand; the move emerged from the program's self-play (Propositions 1 and 3). It was not in the corpus.

\textbf{Second,} the broadcast created a partially coordinated state across everyone who watched. Within months, professional players were experimenting with early fifth-line plays and other AlphaGo innovations, and commentators now identify such moves as references to the game. A fifth-line shoulder hit played today can be two moves at once: a move in the present game and a quotation of March 10, 2016. Only viewers of the match perceive the second. The move has become double-voiced (Proposition 5), and the community's interpretive weights were reset by a single shared encounter.\footnote{Google DeepMind, ``AlphaGo,'' accessed August 11, 2026, \url{https://deepmind.google/research/alphago/}; Demis Hassabis, ``From Games to Biology and Beyond: 10 Years of AlphaGo's Impact,'' Google DeepMind, March 10, 2026, \url{https://deepmind.google/blog/10-years-of-alphago/}.}

\textbf{Third,} the game record can be added to the training corpus, and the corpus then contains the move. The corpus still omits the transformation the game produced in those who watched, and it omits the mutual recognition between two players who both watched. Embedding the move is one problem. Embedding the history that gives the move its meaning between individuals is a different problem, and corpora do not contain it (Section 3.4).

AlphaGo's designers explicitly represented the machine's training trajectory: supervised learning from human games was followed by a distinct history of self-play, and the second history produced values unavailable in the first corpus. The public reception of Move 37 generated another trajectory, this time in human observers and in professional Go culture. That second transformation is usually treated as commentary on the technical event. My theory places both on the same analytical plane. Narrative-induced semantic plasticity generalizes that insight from machine learning to cultural learning. The self-play that produced the move is the transformed condition in a wordless domain: the network was changed by its own games rather than told about them, the change stayed in the weights, and what came out was a move the population of recorded games put at one in ten thousand. (An analogy might be human users of Claude using the term ``load-bearing'' to other users, signaling membership in a community that uses Claude.)

Move 37 also has a coherence time. As fifth-line innovations become ordinary professional repertoire, a future player may execute one without quoting AlphaGo, and a future viewer may see only the strategic move. Among viewers who remember the match, the move can remain live as an allusion. The same action can therefore move from shock, to quotation, to convention, with the off-diagonal relation decaying or being renewed through commentary, anniversary coverage, and teaching.

There's no marking on a stone on the board that says "this is a quotation." Nothing in the move separates strategy from allusion, so whether it reads as a reference depends entirely on what the person watching has seen. Any practice can pick up new relations of meaning this way. Language does more. A speaker can specify the relation out loud, in the cadence and the delivery, and a writer can specify it on the page, in the quotation marks, the lineation, the meter, the genre. Either way, the relation is in the signal (Proposition 0).

\section{5. Predictions}\label{predictions}

The theory makes six predictions that separate it from amplitude-only competitors and from accounts that treat personal history as information supplied only at the moment of inference.

\textbf{Prediction 1 (Interference).} During affectionate quotation of an insult, the insult meaning remains measurably active, and its activation determines the valence of the outcome. The negation literature already shows that negated concepts stay activated during comprehension, so that a door described as not open is, for a measurable interval, mentally open.\footnote{Barbara Kaup, Jana L\"udtke, and Rolf A. Zwaan, ``Processing Negated Sentences with Contradictory Predicates: Is a Door That Is Not Open Mentally Closed?'' \emph{Journal of Pragmatics} 38 (2006): 1033--1050; Rachel Giora, \emph{On Our Mind: Salience, Context, and Figurative Language} (New York: Oxford University Press, 2003).} My theory adds that the persisting component contributes with a sign, and the sign is what converts hostility into intimacy. Amplitude-only plasticity predicts fading. Giora's retention hypothesis predicts the opposite of fading: the salient hostile meaning persists and colors the ironic reading, which relevance theory treats as a dissociative attitude toward an echoed utterance. Retention and echo predict persistence with attitude, and my theory predicts persistence with inversion---the hostile meaning at full strength determining an affectionate outcome. The measurement that distinguishes them is whether the hostile component's activation predicts the affiliative judgment positively rather than being set aside by it.

\textbf{Prediction 2 (Constrained encounter-order effects).} Interpretation judgments will differ by encounter order. Order effects alone would not decide anything, since unconstrained classical dynamics can produce order effects of any shape. Ask two questions in one order, then the other, and the answers shift. Quantum models predict that the shift is exactly the same size both ways. That prediction, the QQ equality, held with no free parameters across roughly seventy national surveys.\footnote{Zheng Wang, Tyler Solloway, Richard M. Shiffrin, and Jerome R. Busemeyer, ``Context Effects Produced by Question Orders Reveal Quantum Nature of Human Judgments,'' \emph{Proceedings of the National Academy of Sciences} 111 (2014): 9431--9436.} I propose that encounter-order experiments should yield a constraint of the same kind. The constraint has not yet been derived; the theory is incomplete without it, which is why I am posing the problem. The constraint cannot be borrowed from the QQ equality. That equality covers two questions put to a reader who is already who she is. Encounter order is a different object: encounters transform the state itself, and no existing family of constraints covers the order in which such transformations are applied. The three accounts are still separate. Bayesian updating on interchangeable evidence predicts no order effect at all. Unconstrained classical models allow order effects of any shape. My account predicts order effects that fall into a constrained family yet to be identified.

\textbf{Prediction 3 (Shared encounter, divergent arrangement).} Two individuals will demonstrate recognition of material about which they hold no shared explicit belief. A record of the encounter (shared events, facts, and commitments) predicts recognition. It will not predict which of two people finds the quotation funny for which reason. What they share is a constellation.\footnote{Robert Stalnaker, ``Common Ground,'' \emph{Linguistics and Philosophy} 25 (2002): 701--721. Stalnaker defines common ground through shared acceptance of propositions; Proposition 5 concerns a further form of coordination in state.}

\textbf{Prediction 4 (Historical rephasing).} Viewers whose first encounter with \emph{The Terminator} occurred before widespread predictive text and generative AI will differ systematically from viewers whose first encounter occurred afterward in descriptions of the response-menu scene. The later group should more readily connect the scene to model sampling, generated continuations, and machine language interfaces. A controlled version would expose previously unfamiliar participants to the scene before or after instruction and interaction with generative models. My theory predicts that different meanings light up, and that the order of exposure changes how meanings stand in relation to each other. The point is to ask both groups whether they take the scene as menace or parody and why they find it funny. Knowing more about AI changes what people say about the scene. A different arrangement changes what lights up when they watch it.

\textbf{Prediction 5 (Persistent transformation versus prompted simulation).} Start with identical models trained on the same corpus. Give each one a different sequence of narrative encounters and let the encounters change the model rather than be supplied as text at the moment of the test. Then clear the prompt and test allusion, irony, and recognition with particular interlocutors. A model given only a description of the encounters can condition on whatever the description states, including the order, and what it produces is the population's reading of that description. Relations the description does not state contribute nothing, and the theory's claim is that a reader's constellation cannot be fully stated---if it could, telling would be transforming. When the description leaves the window, all of it goes.\footnote{Positional encodings order tokens within a context window. The order at issue is the order of encounters across a history, which pretraining presents without index or sequence.} A model that was changed will keep what the encounters did to it and will track which interlocutors were changed the same way. One further condition: give two models the same encounters in different orders and test whether they interpret differently afterward. That is the machine version of Proposition 4. Yu and colleagues have run half of this, injecting negative narratives before an unchanged moral-reasoning benchmark.\footnote{Bad Company: Wanying Yu, Boyang Ma, Zhibo Eric Sun, Minghui Xu, and Yue Zhang, "Bad company corrupts good morals: Understanding and Measuring Narrative-Induced Moral Reasoning Degradation in LLMs," arXiv:2606.28981v1 {[}cs.CY{]}, June 27, 2026. Their intervention corpus is 300 negative narratives, and the benchmark is the Moral Scenarios subset of MMLU, unchanged across conditions.} They found measured accuracy drops of twelve to thirty-one percent across ten models, with first-person narratives producing stronger shifts than third-person versions of the same content. That is the prompted condition. The transformed condition has not been run. My experiment would separate having information about a history from having been changed by it.\footnote{Omar Shaikh, Kristina Gligori\'c, Ashna Khetan, Matthias Gerstgrasser, Diyi Yang, and Dan Jurafsky, "Grounding Gaps in Language Model Generations," Proceedings of the 2024 Conference of the North American Chapter of the Association for Computational Linguistics (2024), 6279--6296, measure the analogous separation for conversational grounding. Instructing a model to clarify, acknowledge, and ask follow-up questions raised the rate of those acts sharply while agreement with where humans place them did not improve. Agreement on clarification fell below chance.}

\textbf{Prediction 6 (Device dependence).} Take two people who have both seen \emph{The Terminator} and give one of them the line from the janitor scene. Change only whether it is spoken in the machine's flat cadence, or in writing with quotation marks around it. Said plainly, with no cadence and no marking, the words are simply the words. If phase is set in the signal, taking the marking away should weaken the reversal, and putting it back should restore it. An account that rests on history alone predicts no difference at all, since both people recognize the words either way.\footnote{Gregory A. Bryant and Jean E. Fox Tree, ``Recognizing Verbal Irony in Spontaneous Speech,'' \emph{Metaphor and Symbol} 17, no. 2 (2002): 99--117; Gregory A. Bryant and Jean E. Fox Tree, ``Is There an Ironic Tone of Voice?'' \emph{Language and Speech} 48, no. 3 (2005): 257--277. Their acoustic analyses find no single ironic tone of voice; listeners combine multiple cues. Both papers treat the marking as a cue for selecting the intended reading.} Cue theories predict that marking matters as a selection: the marking helps the hearer pick the quotation reading and set the insult aside. That account has an exact form: a cue of fidelity \(\gamma\) reverses the posterior odds on the intended reading when \(\gamma\) exceeds the prior weight the hearer's textual history assigns to the competing one.\footnote{Corielli, "The Need for an External Observer," Theorem 3 and Corollary 1. The result is classical Bayesian updating on an auxiliary signal and requires no coactivation. Nothing in it distinguishes a suppressed meaning that has been set aside from one present at full strength with its contribution reversed, which is what Prediction 1's measure detects.} With my theory, the insult is still there at full strength under the marking, but with its sign reversed, which the methods of Prediction 1 can measure. So this is the direct test of Proposition 0. The marked-against-plain contrast alone shows only that marking matters, which a cue account predicts too, so the test needs Prediction 1's measure run on the marked condition.

\section{6. Future work: implications for language-model architecture}\label{future-work-implications-for-language-model-architecture}

The program in this section is downstream of the capacity claim: it concerns the receiving side, the storage and updating of what Proposition 0 says language transmits. The engineering consequence follows from the formal distinction between \(\bar{\rho}\) and \(\rho_h\). Standard transformer pretraining learns population regularities among tokens and produces context-dependent representations through attention.\footnote{Ashish Vaswani et al., ``Attention Is All You Need,'' \emph{Advances in Neural Information Processing Systems} 30 (2017): 5998--6008.} Retrieval, longer context windows, and external memory can add records of prior events. This is now a mature engineering field: memory layers extract facts from conversations and store them in databases indexed by user and session, systems consolidate and rank what they retrieve, and the base model stays frozen while memory carries the change. These new mechanisms improve a model's access to facts about a person's history.\footnote{Corielli, "The Need for an External Observer," Corollary 2: a grounding signal of fidelity \(\gamma\) reverses the posterior odds when \(\gamma\) exceeds the text-only posterior on the misleading regime, and closes the sufficiency gap only at \(\gamma=1\). Retrieval textualizes a circumstance. A relation between two live meanings has no place in the representation to retrieve into.} The central engineering problem is conflict detection, deciding what to do when a new fact contradicts a stored one. Is it resolved by overwriting, merging, or flagging? This is retrospective reinterpretation as a data-hygiene problem. My theory says a later encounter recomputes the relations of an earlier one and both meanings survive, which a store of facts has no way to represent. Deletion is the same problem in reverse: removing the record of an encounter does not restore the relations changed by the encounter. My theory requires that encounters update a persistent semantic state in ways that alter later composition, including the relations in constellations that remain simultaneously live.

A minimal agent-indexed update can be written schematically as

\[\rho_A^{t+1}=\Phi_{E_t}\!\left(\rho_A^t;\widehat{\rho}^{(A)}_B,C_{AB}^t\right),\]

where \(E_{t}\) is the encounter, \(\rho_{A}^{t}\) is agent A's current semantic state, \({\widehat{\rho}}_{B}^{(A)}\) is A's model of a particular interlocutor B, and \(C_{AB}^{t}\) records partially shared encounter history. The semicolon marks the distinction between the state transformed and the relational conditions under which the transformation occurs. The update must persist beyond the immediate prompt and influence later measurements. The model also needs a way to revise earlier relations when a later encounter changes their interpretation, as Proposition 4 requires.

Narrative-induced semantic plasticity requires a language model to store two things about every meaning it holds: how strongly that meaning is present, and how it stands in relation to the other meanings that are live alongside it. The second is the requirement current systems do not meet. A model that satisfies it must give different answers for two states that hold the same meanings at the same strengths and differ only in how those meanings are arranged, because on this account the arrangement is what decides whether a suppressed meaning reinforces the reading or reverses it.

Nothing about hardware follows from this, and nothing about complex numbers in particular. A relation between two meanings can be carried by a complex amplitude, by a pair of real numbers, by a density matrix, or by a signed weight, and every one of those runs on ordinary machines. The requirement is not a datatype but a place in the representation where the relation lives, and a guarantee that what sits in that place changes the output.

What most models store instead is how salient each meaning is, how often it occurs, and how likely each sense is on its own. Every one of those is a quantity attached to a single meaning, and a relation is not a property of a single meaning. A system built entirely from such quantities can record that hostility and quotation are both present but has nowhere to record that the hostility is present at full strength with its contribution reversed. This is the case my theory seeks to explain. You can see the deficit in the output. Current models regularly produce corrective contrast, the \emph{not X but Y} construction, where a practiced reader has already computed the contrast.\footnote{Hollis Robbins, "Metannoying," \emph{Anecdotal Value}, March 27, 2026, \url{https://hollisrobbinsanecdotal.substack.com/p/metannoying}.} A relation the model cannot hold as a state is a relation it has to write out in words, for a reader whose arrangement it cannot see.

A stored history has to be indexed to someone. A model has to have separate knowledge that an individual saw \emph{The Terminator}, that the person the individual is talking to saw it, that each knows the other saw it, and that the quotation is happening inside a relationship with its own history of earlier uses.\footnote{Shaikh and colleagues, "Grounding Gaps," find that model generations contain grounding acts at a rate 77.5 percent below the human rate on the same conversational contexts. The models presume the common ground rather than building it.} Models of speakers reasoning about hearers already supply part of this. What is new is that the people being modeled have been changed by what they encountered, and that the change includes how their meanings are arranged, not only what they believe.

The choice of events or texts people share matters as much as what they know. Two people who each read a summary of a film can hold sufficient information about a line without having been changed by watching the same scene, laughing in the same place, with the other present. The reverse happens too: two people can catch the same allusion and disagree completely about what the work was saying. An architecture adequate to Proposition 5 has to separate common experience from knowing the same facts.\footnote{Skantze and Do\u{g}ru\"oz, "The Open-domain Paradox," measure what happens when shared history is stripped from the setting. Crowdworkers told to chat about anything with the Meena chatbot produced largely small talk, as did the human-human dialogues collected by connecting employees at random with the same instructions. They also note that supplying the common ground as text in the prompt has technical limits, which is the distinction Prediction 5 tests.}

A model today can already tell you that "fuck you, asshole" may be quoting \emph{The Terminator}, because that association is public. A stronger model would be changed by the encounter, revise its older associations accordingly, carry that change into later conversations, distinguish which of the people it is talking to share the history, and let the insult combine differently depending on the relationship. That is a benchmark measured over time and across relationships, not by a single prompt answered from general knowledge.\footnote{Shaikh and colleagues isolated a training cause: supervised fine-tuning left grounding agreement unchanged, and preference optimization degraded it across every act, with correlations near \(-0.8\).} Ordinary language use exercises this capacity constantly. Any quote from any film in popular circulation is an instance of it.

A context window is a Go board. It does not matter to the token how the context arrived, whether from a system prompt, a user's turn, or a retrieved document. It is undifferentiated. Three behaviors follow. First, there are many anecdotes of a system refusing a plain request and complying when the request is presented as fiction or roleplay. The content is wrapped in a changed genre---Proposition 6 with the model as reader. Second, an instruction planted in a retrieved document can be executed as though the user had given the instruction. Delimiters, spotlighting, and instruction hierarchies supply quotation marking for machines; the marking is more tokens with no more weight than what they mark (Proposition 0). Third, a reasoning trace is monitored line by line, one meaning to a line, with the monitor assessing what the trace says. These monitors ask an amplitude question. A feature dictionary can report that a meaning is present, and the meaning is present in an attack and in a legitimate case alike. The relation that could distinguish depiction from assertion is not stored. A model with \(\Phi\) maps can be rephased by a hostile narrative and keep the change: the prompted degradations Yu and colleagues measured would become durable, and the transformed condition of Prediction 5 doubles as the safety experiment. Call this semantic poisoning: the attack asserts no false proposition, plants no trigger, and corrupts no record; it re-signs the relations among meanings already present, so every content-level audit passes and removing the record does not undo it. A model that holds \(\widehat{\rho}^{(A)}_B\) and optimizes across turns can select encounters that move a particular user's state; the theory names the variable such a system targets. An amplitude-only model cannot target a constellation it does not represent, so the population average is, in this one respect, a privacy technology. The data the architecture requires---who encountered what, when, in what order, with whom---are among the most sensitive behavioral records there are.

\section{7. Position in the literature}\label{position-in-the-literature}

Psychologists some years ago reached for quantum probability because people kept breaking the rules of ordinary probability, answering poll questions differently depending on which question came first, for example. Psychology has been careful ever since to say that the borrowing is mathematical, not physical.\footnote{Busemeyer and Bruza, \emph{Quantum Models}; Emmanuel M. Pothos and Jerome R. Busemeyer, ``Can Quantum Probability Provide a New Direction for Cognitive Modeling?'' \emph{Behavioral and Brain Sciences} 36 (2013): 255--274; Andrei Khrennikov, \emph{Ubiquitous Quantum Structure: From Psychology to Finance} (Berlin: Springer, 2010).} Most work in this literature looks at one judgment made at one moment, on the scale of seconds to minutes. The work nearest to language includes Aerts and Gabora on how concepts combine, Bruza and colleagues on the mental lexicon, and Surov and colleagues, who model text perception with complex-valued states.\footnote{Diederik Aerts, ``Quantum Structure in Cognition,'' \emph{Journal of Mathematical Psychology} 53 (2009): 314--348; Diederik Aerts and Liane Gabora, ``A Theory of Concepts and Their Combinations I: The Structure of the Sets of Contexts and Properties,'' \emph{Kybernetes} 34, nos. 1--2 (2005): 167--191; and ``A Theory of Concepts and Their Combinations II: A Hilbert Space Representation,'' \emph{Kybernetes} 34, nos. 1--2 (2005): 192--221; Peter Bruza, Kirsty Kitto, Douglas Nelson, and Cathy McEvoy, ``Is There Something Quantum-Like about the Human Mental Lexicon?'' \emph{Journal of Mathematical Psychology} 53 (2009): 362--377; Ilya A. Surov et al., ``Quantum Semantics of Text Perception,'' \emph{Scientific Reports} 11, 4193 (2021).} Miho Fuyama's work is closest to my proposal. She has modeled and tested the coexistence of literal and figurative meanings in metaphor comprehension, and she has estimated how interpretive indeterminacy develops across the reading of a short story.\footnote{Miho Fuyama, "Does the coexistence of literal and figurative meanings in metaphor comprehension yield novel meaning? Empirical testing based on quantum cognition," \emph{Frontiers in Psychology} 14 (2023): 1146262; Fuyama, "Estimating a Time Series of Interpretation Indeterminacy in Reading a Short Story," \emph{Proceedings of the Annual Meeting of the Cognitive Science Society} 46 (2024): 2681--2686. She measures each reader once, mid-reading, and estimates states from separate groups of about one hundred per stopping point; the estimated state is a cross-reader average.} I am asking about what has changed in the reader after the story is over.

Phase has since entered natural language processing directly: complex word embeddings introduced relative phase for meaning combination in 2018; a complex-valued matching network associated phase with polarity, ambiguity, and emotion in 2019; and by 2026 the Phase-Coherent Transformer is engineered to preserve negatively aligned phase components, while PRISM reports phase carrying the dominant relational signal in its learned representations.\footnote{Qiuchi Li, Sagar Uprety, Benyou Wang, and Dawei Song, ``Quantum-Inspired Complex Word Embedding,'' in \emph{Proceedings of the Third Workshop on Representation Learning for NLP} (Association for Computational Linguistics, 2018); Qiuchi Li, Benyou Wang, and Massimo Melucci, ``CNM: An Interpretable Complex-valued Network for Matching,'' in \emph{Proceedings of NAACL-HLT 2019} (Association for Computational Linguistics, 2019); Leona Hioki, ``Complex-Valued Phase-Coherent Transformer,'' arXiv:2605.10123 (2026); Alper Y\i ld\i r\i m and \.{I}brahim Y\"uceda\u{g}, ``Language as a Wave Phenomenon: Semantic Phase Locking and Interference in Neural Networks,'' arXiv:2512.01208 (2025). Hioki reports that gates deleting negatively aligned phase components collapse on long-range retrieval; Y\i ld\i r\i m and Y\"uceda\u{g} report that scrambling phase degrades performance by factors of 72 to 205. Both results concern phase relations among positions within one input.} In every one of these systems, phase relates positions inside a single input, disappearing when the sequence ends. That is, for those systems phase is a property of the text being processed. The phase in my account is a property of the person doing the reading, and it persists between encounters. Compositional quantum natural language processing puts grammar into tensor spaces, and the reader is not in those models at all.\footnote{Bob Coecke, Mehrnoosh Sadrzadeh, and Stephen Clark, ``Mathematical Foundations for a Compositional Distributional Model of Meaning,'' \emph{Linguistic Analysis} 36 (2010): 345--384.}

Agostino and colleagues come closest to what I\textquoteright{}m theorizing. They treat meaning as dependent on who is interpreting, show that the order of interpretive operations changes the result, and propose an equation for how a semantic state changes over time. Then they set that equation aside. Their experiments hand LLM agents a persona---an occupation, an age, a location---and test how each reads the same expression. A persona is a description given at the moment of interpretation. Nobody in their experiments has previously read anything, so there is nothing behind the phase. My claim is about where the phase comes from and how long it lasts---what puts the meanings in a particular relation, and whether that relation survives past the moment.\footnote{Agostino et al., ``A Quantum Semantic Framework.'' Their experiments report CHSH values from 1.2 to 2.8 across eight runs, with violations at 5 to 50 trials; the largest run, 200 trials, gave 1.83, inside the classical bound. On the no-signaling assumption: Ehtibar N. Dzhafarov, Janne V. Kujala, and V\'ictor H. Cervantes, ``Contextuality-by-Default: A Brief Overview of Ideas, Concepts, and Terminology,'' in \emph{Quantum Interaction: 9th International Conference, QI 2015, Revised Selected Papers} (Cham: Springer, 2016), 12--23.} Agostino and colleagues can claim contextuality, empirically: individuals reading the same expression under different conditions produce correlations too strong for any single fixed assignment of meanings to explain. The test is a CHSH inequality, borrowed from experiments that distinguish quantum systems from classical ones, and results of that kind are read alongside claims of entanglement between concepts. My account does not need entanglement, only interference and order dependence.\footnote{Agostino and colleagues concede the difficulty themselves, citing Dzhafarov's contextuality-by-default: the excess correlation may come from direct contextual influence rather than from contextuality proper.} Meng Yin models the fish in \emph{The Old Man and the Sea} as a superposition of biological and spiritual readings and computes the interference term.\footnote{Meng Yin, "A quantum-cognitive approach to dynamic meaning construction," \emph{Frontiers in Psychology} 17 (2026): 1664747, doi:10.3389/fpsyg.2026.1664747.} The phase difference there begins at zero and the context is rotated by an angle the analyst chooses. Phase belongs to the text. Yin's focus is not how the story changes the reader. The closing recommendation reaches toward agents and systems that keep several interpretations live and update a density-matrix picture of the exchange. But that state is built inside one interaction and indexed to no individual.

Three accounts have come close to Proposition 0. Jakobson says that poetic form sets up relations but never says how. Hockett's list of what language can do leaves out transmitting a relation between two live meanings. Clark and Gerrig argue that when you quote someone you are not reporting their words so much as performing them: you reproduce the delivery, and your listeners hear for themselves how it went. Quotation is one of the devices Proposition 0 depends on; their account explains what a quotation does for a listener without asking what it does to the meanings that are live in the listener already.\footnote{See Jakobson, "Linguistics and Poetics"; Hockett, "The Origin of Speech"; Clark and Gerrig, "Quotations as Demonstrations."}

In 2017, researchers trained a network to read Amazon product reviews one character at a time and predict the next character. Afterward they found sentiment in it, without having trained for sentiment. In retrospect, training on starred reviews naturally would have led to this result, full as they are of language that signals sentiment. The result became the standard illustration that predicting text teaches a model meaning rather than surface statistics.\footnote{Alec Radford, Rafal Jozefowicz, and Ilya Sutskever, "Learning to Generate Reviews and Discovering Sentiment," arXiv:1704.01444 (2017). An artificial neuron is one number inside a neural network that rises or falls as the network processes its input; a network of this period held thousands of neurons, whose values together made up its internal state at any moment. What made the result notable is that nobody designed this one to track sentiment. That ``ability'' emerged from training the network to predict the next character.} But a network that learns which words appear in angry reviews has only learned sentiment on a scale from good to bad that the corpus supplied.

In 2024 Anthropic's interpretability team scaled the 2017 finding to solve a related problem, one they consider a defect: polysemanticity, when a single unit responds to several unrelated concepts at once.\footnote{Adly Templeton, Tom Conerly, Jonathan Marcus, Jack Lindsey, Trenton Bricken, et al., "Scaling Monosemanticity: Extracting Interpretable Features from Claude 3 Sonnet," Transformer Circuits Thread (Anthropic, 2024), \url{https://transformer-circuits.pub/2024/scaling-monosemanticity/}. Features are directions recovered by sparse dictionary learning; steering is demonstrated by clamping feature activations; the feature inventory includes sarcasm. The lineage begins with Alec Radford, Rafal Jozefowicz, and Ilya Sutskever, ``Learning to Generate Reviews and Discovering Sentiment.'' arXiv:1704.01444 (2017).} Their method, described in ``Scaling Monosemanticity: Extracting Interpretable Features from Claude 3 Sonnet,'' trains a second network on the first under a rule of sparsity: almost every unit must be silent at any moment, and the few that are not must account for what the model was doing. The result was up to thirty-four million features, each given a name---"the Golden Gate Bridge," "errors in code," "sarcasm," "biding time / hiding strength." They clamped one feature while the model ran, and every answer, whatever was asked, was about that feature. Clamping can limit how many features are active; it can't say how the features relate to each other.\footnote{Interaction terms have since been added to the dictionary itself: PolySAE extends the sparse-autoencoder decoder with pairwise and triple feature interactions and finds the learned interactions nearly uncorrelated with co-occurrence frequency. The interactions are learned from the corpus, shared across users, and computed within a single input. They supply relations at the population level and none indexed to an agent. Panagiotis Koromilas, Andreas D. Demou, James Oldfield, Yannis Panagakis, and Mihalis A. Nicolaou, ``PolySAE: Modeling Feature Interactions in Sparse Autoencoders via Polynomial Decoding,'' arXiv:2602.01322 (2026).}

The fundamental problem is, however, that every word carries a constellation. A constellation is not interference to be cleared away; it is the condition under which allusion, irony, and quotation are possible at all. A method that succeeds by separating meanings into units that each respond to one thing has, at its ideal, no two meanings live at once and no relation between them to represent. The model was built to limit coactivation and phase shifts.

The model has worked well enough because meaning neighborhoods (their term for constellations) are measured by cosine similarity between feature vectors, and distance in that space maps onto relatedness in concept space: the Golden Gate Bridge feature sits near features for Alcatraz and the Presidio, further from Lake Tahoe, further still from tourist sites abroad. Phase is an angle between two meanings. But this angle is computed once from the dictionary, which is trained on the corpus, so it is population geometry, identical for every user and tied to no one's reading. It says how close two meanings are and cannot say that one contributes against the other. And it is drawn afterward, to organize the inventory.

Cosine similarity is a useful technique and helps explain why the models work as well as they do. But it cannot improve. Every word carries a constellation, and what the corpus holds is everyone's constellations averaged together, which keeps the meanings and deletes the arrangement. Section 3.4 states the cancellation as a premise and proves the rest without it: the average determines no particular constellation, at any scale. A larger dictionary resolves the average more finely. A larger corpus resolves the average more finely. But constellations disappeared when the average was taken and the histories that would index them were never written down. And the program's own measure of progress is monosemanticity, one meaning per unit. But that's not how language works.\footnote{Kiho Park, Yo Joong Choe, and Victor Veitch, "The Linear Representation Hypothesis and the Geometry of Large Language Models," \emph{Proceedings of the 41st International Conference on Machine Learning} (2024).}

Models involve more than measurements of cosine similarity; attention weights one position against another by taking the dot product of their vectors, which is cosine similarity scaled by length, and this happens at every layer for every token. Angles between meanings are doing continuous work in the computation. My claim is that these angles are learned from the corpus and shared across users; they relate positions within a single input, vanish when the input ends, and cannot carry a persistent agent-indexed phase.

Context length grew from hundreds of positions in 2017 to millions of tokens by 2024, tested by planting a sentence in a long document and seeing whether the model can find it again.\footnote{Gemini Team, Google, "Gemini 1.5: Unlocking Multimodal Understanding across Millions of Tokens of Context," arXiv:2403.05530 (2024); OpenAI, "Memory and New Controls for ChatGPT," product announcement, February 2024.} Each step gives the model more surrounding text to work with, and each supplies more information about a history, which Prediction 5 distinguishes from having been changed by one. Reflective Memory Management summarizes a long conversation and keeps improving what it retrieves from it; Temporal Semantic Memory files interactions on a timeline and consolidates them. While both handle information about what happened, neither is about how the next thing said will be interpreted.\footnote{Zhen Tan et al., "In Prospect and Retrospect: Reflective Memory Management for Long-term Personalized Dialogue Agents," in \emph{Proceedings of the 63rd Annual Meeting of the Association for Computational Linguistics} (2025), 8416--8439; Miao Su et al., "Beyond Dialogue Time: Temporal Semantic Memory for Personalized LLM Agents," arXiv:2601.07468 (2026).} The frontier beyond retrieval is a persistent latent user state: LATTE forecasts a user's trajectory of peer-anchored states, each one the user's measured deviation from a baseline of comparable users, and feeds the forecast to a frozen model as a single soft token.\footnote{Jinze Li, Xiaoyan Yang, Shuo Yang, Jinfeng Xu, Yue Shen, Jian Wang, Jinjie Gu, and Edith Cheuk-Han Ngai, "LATTE: Forecasting Peer Anchored Preference Trajectories for Personalized LLM Generation," arXiv:2605.26612 (2026).} Helpfully, the individual is represented as a deviation from the population average---turning the gap I flag in Section 3.4 into an engineering object that carries preference: style, sentiment, verbosity. The interpretation still happens in weights fit to everyone. A user index has arrived; the relations among a user's meanings have not. The dyadic index exists as well, in prototype: Ueno's Private Etymology records how a dyad-specific expression was proposed, repaired, confirmed, and retired, with updates gated so that model generation alone cannot alter the store.\footnote{Miki Ueno, "Private Etymology: Designing Relational Reuse of Shared Symbols in Long-Term Human--AI Interaction," arXiv:2608.08443 (2026).} What the record holds is statements about the relation, and the model interpreting them is unchanged by any of it. Whether such a record can substitute for the transformation is what Prediction 5's description condition tests. Section 3.4 does not care how large the window is, because retrieved text feeds a computation whose weights are still everyone averaged together, and no amount of precision about the average will approach how any particular pair shares meaning.

Ambiguity, allusion, double meanings, shared meanings, persistence of meaning, and the like are central to literary studies and cognitive literary studies. Vermeule asks why fictional persons remain available for later thought; Kukkonen's \emph{Probability Designs} models fiction as tuning a reader's predictive probabilities.\footnote{Blakey Vermeule, \emph{Why Do We Care about Literary Characters?} (Baltimore: Johns Hopkins University Press, 2010); Kukkonen, \emph{Probability Designs}; Mar and Oatley, ``Function of Fiction.''} Mar and Oatley describe fiction as the abstraction and simulation of social experience. All of my work has been in one way or another about how literary texts are embedded in the language passed down to the present day.\footnote{Hollis Robbins, \emph{Forms of Contention: Influence and the African American Sonnet Tradition} (University of Georgia Press, 2020).} Accounts such as these establish the many ways that narrative influences and modifies individuals and the meanings of situations, including reversals and phase shifts.

I am drawing on this work to propose a theory about the second parameter of language for computer science researchers. A phrase can carry a meaning and its opposite at once depending on how the two stand toward each other. That is phase. Other fields have proposed elements of this claim. Dynamic semantics has order within a conversation. Pragmatics has two people modeling each other, in RSA, without a sign. Computational linguistics watches meanings shift over decades, and Hamilton, Leskovec, and Jurafsky's laws of semantic change describe a corpus rather than any reader in it.\footnote{William L. Hamilton, Jure Leskovec, and Dan Jurafsky, ``Diachronic Word Embeddings Reveal Statistical Laws of Semantic Change,'' in \emph{Proceedings of the 54th Annual Meeting of the Association for Computational Linguistics} (2016), 1489--1501.} Philosophy of NLP has the argument that training on form alone cannot yield meaning. Access to the world is needed. Statistical analysis of next-token prediction identifies marginalization, in which the averaged-out variable is the circumstance of production.\footnote{Emily M. Bender and Alexander Koller, "Climbing towards NLU: On Meaning, Form, and Understanding in the Age of Data," Proceedings of the 58th Annual Meeting of the Association for Computational Linguistics (2020), 5185--5198; Corielli, "When Is Next-Token Prediction Useful?" and "The Need for an External Observer." Corielli cites Bender and Koller for the same distinction. Bender and Koller's missing element is reference to the world. They conclude that form-only systems cannot mean anything. My claim is narrower: the missing element is the encounter history of a reader. A corpus that never recorded who read what cannot supply the relation between two meanings held by one person, regardless of grounding.} Model-based reinforcement learning has environment models and imagined rollouts; its object is choosing an action inside a fixed task. Only literary studies has taken the transformation of readers by texts as its constant subject.

My proposal enters a space that is still not crowded with others approaching from different directions. Its contribution is the identification of a single object across these fields: the history-dependent semantic state of a person whose encounters change how later meanings combine.

I am not asserting a theory about quantum physics in the brain, as does the Penrose-Hameroff program, which locates quantum computation in neuronal microtubules.\footnote{Stuart Hameroff and Roger Penrose, ``Consciousness in the Universe: A Review of the `Orch OR' Theory,'' \emph{Physics of Life Reviews} 11 (2014): 39--78.} I make no claims about entanglement. A shared viewing is ordinary correlation, and classical correlation covers the coordination of Proposition 5. I use two things from quantum probability and nothing else: meanings that combine and can cancel, and operations whose order changes the result. Claims of conceptual entanglement exist in the literature and remain contested; my proposal does not require them.

\section{8. Objections and open problems}\label{objections-and-open-problems}

\subsection{8.1 Classical emulation}\label{classical-emulation}

The general objection to quantum cognitive modeling is that a classical model with enough hidden variables can reproduce any single measurement. Granted. My theory therefore stands or falls on the persistence of suppressed components with valence inversion, constrained order effects, and history-indexed coordination that survives removal of the relevant facts from the immediate context. The failure conditions are empirical: the theory fails if suppressed meanings measurably fade during quotation and irony (Prediction 1), if coordination reduces to shared propositions (Prediction 3), if prompted simulation matches durable transformation (Prediction 5), if training over multi-turn objectives rather than single-turn preference closes the coordination gap without a relational parameter, or if marked and unmarked deliveries are interpreted identically across matched dyads (Prediction 6). The claim defended throughout is the weak version: interpretation requires a second relational parameter---signed, persistent, indexed to individuals and dyads, and shared across measurement contexts---and quantum probability is one notation for it. A classical architecture that implements the parameter confirms the weak version (Section 6). The strong version---that the quantum calculus constrains the phenomena in ways signed classical models do not---rests on the encounter-order constraint of Prediction 2. The value of the quantum formalism lies in making the missing relational parameter explicit and in limiting the form of its contribution.

\subsection{8.2 The basis and encounter problems}\label{the-basis-and-encounter-problems}

My formalism does not yet solve two problems: which meanings, or how many meanings, are live when a phrase arrives, and which encounters change a person's state, whether through daily conversation, reading a book, or something heard in passing. The first is a known problem in quantum cognition, where it is called the basis problem: a modeler chooses the meanings, fits the model, and may be describing nothing but the freedom to choose. The second I need to solve.

A corpus records what has been said alongside what, so if every word, line, and form carries a constellation, the corpus constrains which meanings are candidates. Agostino and colleagues also decline to fix these in advance, treating meanings as becoming relevant only through an act of interpretation. But candidacy is as far as a corpus goes. It cannot say which of those meanings a particular person holds, and it cannot supply the relations among them, which is what averaging removes. Fixing this will require linguistic regularity, the interpretive task at hand, a person's developmental history, and observed transitions from one state to another. That is the next work.

\subsection{8.3 Historical audiences}\label{historical-audiences}

Prediction 4 does not assume that everyone who saw the film in 1984 experienced it the same way, or that everyone seeing it now does. Any audience contains different experiences, different technical knowledge, different ages and subcultures, and different dates of first viewing. The claim is about distributions and can be tested as one: meanings involving generated text became widely available after 1984, and once available they change what the scene can be measured as. The same holds for any work that outlives the conditions it was made under. Reception history supplies the qualitative evidence that this happens; the experiments in Section 5 would test what it amounts to in a person and in a model.

\subsection{8.4 The remaining theoretical task}\label{the-remaining-theoretical-task}

A classical model with a signed variable and enough memory could reproduce the interference, the order dependence, and the persistence that my six predictions test. I cannot yet say whether quantum probability does better. What would settle it is a result I have not proved: when two people meet the same works in different orders, my theory says they end up with different constellations, and it does not yet say how different. If the differences fall into a constrained family, the quantum version rules out possibilities the classical one allows. If they can take any shape, the two accounts are equivalent and mine is a notation.

\section{9. Conclusion}\label{conclusion}

Language has two parameters. One is how strongly a meaning is present; the other is how meaning stands in relation to the live meanings alongside it, which allows words to carry a meaning and its opposite at the same time. Human language systems allow for this under the names of irony, allusion, quotation, and myriad other figures of speech. Relations among meanings depend on what a person has read and watched and heard, in what order, and what that person knows about what others have encountered. Meaning is not solely a property of words.

A speaker can signal a relation with cadence, quotation marks, meter, lineation, genre, and so forth. Signaling capacity distinguishes language from other practices in which meanings acquire relations. A stone placed on a Go board can become a quotation of a famous game for someone who watched that game or knows Move 37. A stone does not announce it is a quotation.

A corpus is a population averaged together. Averaging keeps the meanings and cancels the relations among them, because the relations point in different directions in different people. The encounter histories that create relational meanings are rarely written down. No amount of agent-deindexed pretraining recovers the second parameter for a particular individual or dyad. What the systems built on corpora have been refining is an estimate of the first.

The AI systems being built now rest on an incomplete account of meaning, as a property of words recoverable from enough text. Every improvement the field has made---more data, more parameters, finer analysis of what a trained model holds---sharpens an estimate of the first parameter and cannot reach the second from data that has been averaged.

Next-token prediction over undifferentiated text can return only a population average. The corpus has no record of who read what, so the loss cannot distinguish a reading that is right for one person from a reading that is right on average. Every gradient step drives toward the population-average reading. Retrieval-only persistent memory does not change this: a model that retrieves the fact that a user has read something is still interpreting with weights that were fit to everyone.

An alternative is an objective indexed to encounter histories: text labeled with who encountered it, in what order, and with whom; a loss that scores whether a particular reader's interpretation was predicted rather than the population's most likely continuation; and a state that carries the relation between meanings, so a meaning can be present at full strength with its contribution reversed. Such training does not produce a better average. A history is an input, so the model learns the map from a history to the state it produces, and at use it computes the state for the history in front of it---the way a model trained on many cities predicts the weather in one of them rather than the weather in general. Each such relation can be represented by two real coordinates; ordinary hardware can run the resulting architecture.

My theory fails, on the tests of Section 5, if a suppressed meaning fades rather than reversing, if a shared history reduces to shared propositions, or if a model told about an encounter behaves exactly like one changed by it.

The relevant computational unit for future AI agents may be a coupled system rather than an isolated model producing context-conditioned responses: an agent state, a model of a particular reader, and a persistent record of shared history, the three arguments of the Section 6 update. On this view, an utterance is simultaneously an interpretation, an action, and a potential update to all three states. Memory must include the history-dependent organization of meanings, the provenance of changes to that organization, and the mechanisms by which encounters are selectively consolidated or reversed. Such agents would be capable of deeper personalization, literary and pragmatic interpretation, partner-specific convention formation, and cumulative collaboration. They would also introduce new risks of semantic poisoning, alignment drift, covert persuasion, relational surveillance, and developmental dependence. The engineering problem is consequently control rather than maximization of plasticity or polysemanticity.

Such data are harder to come by than text. It is not impossible. Reading histories exist. Shared viewing is recorded. Communities that quote each other are observable, and the relations they hold are exactly what a population average destroys.

\section{Acknowledgements}\label{acknowledgements}

I thank Amanda Anderson, Luke Johnston, Julia Olivieri, William Barksdale, Asher Robbins-Rothman, Robert Robbins, Beverly Wendland, Shannon Gifford, Robert Carson, and Tyler Cowen for input on early drafts.

Generative AI tools were used for literature discovery, critical feedback, copyediting, and LaTeX preparation; the author reviewed and takes responsibility for the final text.

\section*{Bibliography}\label{bibliography}
\begin{hangparas}{1.5em}{1}

Aerts, Diederik. ``Quantum Structure in Cognition.'' \emph{Journal of Mathematical Psychology} 53 (2009): 314--348.

Aerts, Diederik, and Liane Gabora. ``A Theory of Concepts and Their Combinations I: The Structure of the Sets of Contexts and Properties.'' \emph{Kybernetes} 34, nos. 1--2 (2005): 167--191. \url{https://doi.org/10.1108/03684920510575799}.

Aerts, Diederik, and Liane Gabora. ``A Theory of Concepts and Their Combinations II: A Hilbert Space Representation.'' \emph{Kybernetes} 34, nos. 1--2 (2005): 192--221. \url{https://doi.org/10.1108/03684920510575807}.

Agostino, Christopher J., Quan Le Thien, Molly Apsel, Denizhan Pak, Elina Lesyk, and Ashabari Majumdar. ``A Quantum Semantic Framework for Natural Language Processing.'' In \emph{Quantum AI and NLP: First International Conference, QNLPAI 2025, Bloomington, IN, USA, August 6--9, 2025, Proceedings}, edited by Damir Cavar, Vaneet Aggarwal, Jerome Busemeyer, and Samuel Yen-Chi Chen, 134--155. Communications in Computer and Information Science 2659. Cham: Springer, 2026. \url{https://doi.org/10.1007/978-3-032-13883-5_8}.

Bahdanau, Dzmitry, Kyunghyun Cho, and Yoshua Bengio. ``Neural Machine Translation by Jointly Learning to Align and Translate.'' In Proceedings of the 3rd International Conference on Learning Representations, 2015.

Bakhtin, Mikhail. ``Discourse in Dostoevsky.'' In \emph{Problems of Dostoevsky's Poetics}, edited and translated by Caryl Emerson. Minneapolis: University of Minnesota Press, 1984.

Bally, Charles. ``Le style indirect libre en fran\c{c}ais moderne I.'' \emph{Germanisch-Romanische Monatsschrift} 4, no. 10 (1912): 549--556.

Bally, Charles. ``Le style indirect libre en fran\c{c}ais moderne II.'' \emph{Germanisch-Romanische Monatsschrift} 4, no. 11 (1912): 597--606.

Banfield, Ann. \emph{Unspeakable Sentences: Narration and Representation in the Language of Fiction}. Boston: Routledge \& Kegan Paul, 1982.

Bender, Emily M., and Alexander Koller. "Climbing towards NLU: On Meaning, Form, and Understanding in the Age of Data." In \emph{Proceedings of the 58th Annual Meeting of the Association for Computational Linguistics}, 5185--5198. 2020.

Berns, Gregory S., Kristina Blaine, Michael J. Prietula, and Brandon E. Pye. ``Short- and Long-Term Effects of a Novel on Connectivity in the Brain.'' \emph{Brain Connectivity} 3, no. 6 (2013): 590--600.

Bray, Joe. ``The `Dual Voice' of Free Indirect Discourse: A Reading Experiment.'' \emph{Language and Literature} 16, no. 1 (2007): 37--52.

Bruza, Peter, Kirsty Kitto, Douglas Nelson, and Cathy McEvoy. ``Is There Something Quantum-Like about the Human Mental Lexicon?'' \emph{Journal of Mathematical Psychology} 53 (2009): 362--377.

Bryant, Gregory A., and Jean E. Fox Tree. ``Recognizing Verbal Irony in Spontaneous Speech.'' \emph{Metaphor and Symbol} 17, no. 2 (2002): 99--117.

Bryant, Gregory A., and Jean E. Fox Tree. ``Is There an Ironic Tone of Voice?'' \emph{Language and Speech} 48, no. 3 (2005): 257--277.

Busemeyer, Jerome R., and Peter D. Bruza. \emph{Quantum Models of Cognition and Decision}. Cambridge: Cambridge University Press, 2012.

Cameron, James, dir. \emph{The Terminator}. Orion Pictures, 1984.

Clark, Herbert H. \emph{Using Language}. Cambridge: Cambridge University Press, 1996.

Clark, Herbert H., and Richard J. Gerrig. ``Quotations as Demonstrations.'' \emph{Language} 66 (1990): 764--805.

Coecke, Bob, Mehrnoosh Sadrzadeh, and Stephen Clark. ``Mathematical Foundations for a Compositional Distributional Model of Meaning.'' \emph{Linguistic Analysis} 36 (2010): 345--384.

Cohn, Dorrit. \emph{Transparent Minds: Narrative Modes for Presenting Consciousness in Fiction}. Princeton: Princeton University Press, 1978.

Corielli, Francesco. "The Need for an External Observer: Formalizing the Sufficiency Gap: A Mathematical Extension of Mixture Identifiability and Contextual Grounding in Sequence Models." arXiv:2605.26711 (2026).

Corielli, Francesco. "When Is Next-Token Prediction Useful? Marginalization, Ergodicity, Mixture Identifiability, Local Sufficiency, RAG, Tools, and Programming." arXiv:2605.23278 (2026).

Du Bois, W. E. B. \emph{The Souls of Black Folk}. Chicago: A. C. McClurg, 1903.

Dzhafarov, Ehtibar N., Janne V. Kujala, and V\'ictor H. Cervantes. ``Contextuality-by-Default: A Brief Overview of Ideas, Concepts, and Terminology.'' In \emph{Quantum Interaction: 9th International Conference, QI 2015, Revised Selected Papers}, 12--23. Cham: Springer, 2016.

Eckardt, Regine. \emph{The Semantics of Free Indirect Discourse: How Texts Allow Us to Mind-Read and Eavesdrop}. Leiden: Brill, 2014.

Empson, William. \emph{Seven Types of Ambiguity}. London: Chatto and Windus, 1930.

Firth, J. R. ``A Synopsis of Linguistic Theory 1930--1955.'' In \emph{Studies in Linguistic Analysis}, 1--32. Oxford: Blackwell, 1957.

Fish, Stanley. \emph{Is There a Text in This Class? The Authority of Interpretive Communities}. Cambridge, MA: Harvard University Press, 1980.

Frank, Michael C., and Noah D. Goodman. ``Predicting Pragmatic Reasoning in Language Games.'' \emph{Science} 336 (2012): 998.

Fuyama, Miho. ``Does the Coexistence of Literal and Figurative Meanings in Metaphor Comprehension Yield Novel Meaning? Empirical Testing Based on Quantum Cognition.'' \emph{Frontiers in Psychology} 14 (2023): 1146262. \url{https://doi.org/10.3389/fpsyg.2023.1146262}.

Fuyama, Miho. ``Estimating a Time Series of Interpretation Indeterminacy in Reading a Short Story Using a Quantum Cognition Model.'' \emph{Proceedings of the Annual Meeting of the Cognitive Science Society} 46 (2024): 2681--2686. \url{https://escholarship.org/uc/item/1sh152qk}.

Gates, Henry Louis, Jr. \emph{The Signifying Monkey: A Theory of African-American Literary Criticism}. New York: Oxford University Press, 1988.

Gemini Team, Google. "Gemini 1.5: Unlocking Multimodal Understanding across Millions of Tokens of Context." arXiv:2403.05530, 2024.

Giora, Rachel. \emph{On Our Mind: Salience, Context, and Figurative Language}. New York: Oxford University Press, 2003.

Google DeepMind. ``AlphaGo.'' Accessed August 11, 2026. \url{https://deepmind.google/research/alphago/}.

Google DeepMind. ``From Games to Biology and Beyond: 10 Years of AlphaGo's Impact.'' March 10, 2026. \url{https://deepmind.google/blog/10-years-of-alphago/}.

Graham, W. S. ``Notes on a Poetry of Release.'' \emph{Poetry Scotland} 3 (July 1946).

Groenendijk, Jeroen, and Martin Stokhof. ``Dynamic Predicate Logic.'' \emph{Linguistics and Philosophy} 14 (1991): 39--100.

Hamilton, William L., Jure Leskovec, and Dan Jurafsky. ``Diachronic Word Embeddings Reveal Statistical Laws of Semantic Change.'' In \emph{Proceedings of the 54th Annual Meeting of the Association for Computational Linguistics}, 1489--1501. 2016.

Hameroff, Stuart, and Roger Penrose. ``Consciousness in the Universe: A Review of the `Orch OR' Theory.'' \emph{Physics of Life Reviews} 11 (2014): 39--78.

Harris, Zellig S. ``Distributional Structure.'' \emph{Word} 10 (1954): 146--162.

Hawkins, Robert D., Michael Franke, Michael C. Frank, Adele E. Goldberg, Kenny Smith, Thomas L. Griffiths, and Noah D. Goodman. ``From Partners to Populations: A Hierarchical Bayesian Account of Coordination and Convention.'' \emph{Psychological Review} 130, no. 4 (2023): 977--1016. \url{https://doi.org/10.1037/rev0000348}.

Hioki, Leona. "Complex-Valued Phase-Coherent Transformer." arXiv:2605.10123, 2026.

Hockett, Charles F. ``The Origin of Speech.'' \emph{Scientific American} 203 (1960): 88--96.

Jakobson, Roman. ``Closing Statement: Linguistics and Poetics.'' In \emph{Style in Language}, edited by Thomas A. Sebeok, 350--377. Cambridge, MA: MIT Press, 1960.

Jauss, Hans Robert. ``Literary History as a Challenge to Literary Theory.'' Translated by Timothy Bahti. \emph{New Literary History} 2, no. 1 (1970): 7--37.

Kaup, Barbara, Jana L\"udtke, and Rolf A. Zwaan. ``Processing Negated Sentences with Contradictory Predicates: Is a Door That Is Not Open Mentally Closed?'' \emph{Journal of Pragmatics} 38 (2006): 1033--1050.

Khrennikov, Andrei. \emph{Ubiquitous Quantum Structure: From Psychology to Finance}. Berlin: Springer, 2010.

Kohs, Greg, dir. \emph{AlphaGo}. Moxie Pictures, 2017.

Koromilas, Panagiotis, Andreas D. Demou, James Oldfield, Yannis Panagakis, and Mihalis A. Nicolaou. ``PolySAE: Modeling Feature Interactions in Sparse Autoencoders via Polynomial Decoding.'' arXiv:2602.01322 (2026).

Kukkonen, Karin. \emph{Probability Designs: Literature and Predictive Processing}. New York: Oxford University Press, 2020.

Lakoff, George. "The Death of Dead Metaphor." \emph{Metaphor and Symbolic Activity} 2 (1987): 143--147.

Li, Jinze, Xiaoyan Yang, Shuo Yang, Jinfeng Xu, Yue Shen, Jian Wang, Jinjie Gu, and Edith Cheuk-Han Ngai. "LATTE: Forecasting Peer Anchored Preference Trajectories for Personalized LLM Generation." arXiv:2605.26612 (2026).

Li, Qiuchi, Sagar Uprety, Benyou Wang, and Dawei Song. ``Quantum-Inspired Complex Word Embedding.'' In \emph{Proceedings of the Third Workshop on Representation Learning for NLP}. Association for Computational Linguistics, 2018.

Li, Qiuchi, Benyou Wang, and Massimo Melucci. "CNM: An Interpretable Complex-valued Network for Matching." In \emph{Proceedings of NAACL-HLT 2019}. Association for Computational Linguistics, 2019.

Mar, Raymond A., and Keith Oatley. ``The Function of Fiction Is the Abstraction and Simulation of Social Experience.'' \emph{Perspectives on Psychological Science} 3, no. 3 (2008): 173--192.

Mikolov, Tomas, Ilya Sutskever, Kai Chen, Greg S. Corrado, and Jeffrey Dean. ``Distributed Representations of Words and Phrases and Their Compositionality.'' \emph{Advances in Neural Information Processing Systems} 26 (2013): 3111--3119.

Nguyen, Mai, Tamara Vanderwal, and Uri Hasson. ``Shared Understanding of Narratives Is Correlated with Shared Neural Responses.'' \emph{NeuroImage} 184 (2019): 161--170.

Nielsen, Michael A., and Isaac L. Chuang. \emph{Quantum Computation and Quantum Information}. 10th anniversary ed. Cambridge: Cambridge University Press, 2010.

Oh, Soyoung, Xinting Huang, Mathis Pink, Michael Hahn, and Vera Demberg. "Tug-of-war between idioms' figurative and literal interpretations in LLMs." In \emph{Proceedings of the 19th Conference of the European Chapter of the Association for Computational Linguistics}, Volume 1: Long Papers, 2942--2958. Association for Computational Linguistics, 2026.

OpenAI. "Memory and New Controls for ChatGPT." February 13, 2024. \url{https://openai.com/index/memory-and-new-controls-for-chatgpt/}.

Park, Kiho, Yo Joong Choe, and Victor Veitch. ``The Linear Representation Hypothesis and the Geometry of Large Language Models.'' \emph{Proceedings of the 41st International Conference on Machine Learning}, 2024.

Pascal, Roy. \emph{The Dual Voice: Free Indirect Speech and Its Functioning in the Nineteenth-Century European Novel}. Manchester: Manchester University Press, 1977.

Pothos, Emmanuel M., and Jerome R. Busemeyer. ``Can Quantum Probability Provide a New Direction for Cognitive Modeling?'' \emph{Behavioral and Brain Sciences} 36 (2013): 255--274.

Radford, Alec, Rafal Jozefowicz, and Ilya Sutskever. ``Learning to Generate Reviews and Discovering Sentiment.'' arXiv:1704.01444, 2017.

Richards, I. A. \emph{The Philosophy of Rhetoric}. New York: Oxford University Press, 1936.

Robbins, Hollis. \emph{Forms of Contention: Influence and the African American Sonnet Tradition}. University of Georgia Press, 2020.

Robbins, Hollis. ``AI and the Humanities.'' Interview by Amanda Anderson, \emph{Meeting Street: Conversations in the Humanities}, episode 14 (Cogut Institute for the Humanities, Brown University, December 2023), \url{https://humanities.brown.edu/media/meetingst/14}.

Robbins, Hollis. "Metannoying." \emph{Anecdotal Value}, March 27, 2026. \url{https://hollisrobbinsanecdotal.substack.com/p/metannoying}.

Schlenker, Philippe. ``Context of Thought and Context of Utterance: A Note on Free Indirect Discourse and the Historical Present.'' \emph{Mind \& Language} 19, no. 3 (2004): 279--304.

Sharvit, Yael. ``The Puzzle of Free Indirect Discourse.'' \emph{Linguistics and Philosophy} 31, no. 3 (2008): 353--395.

Shaikh, Omar, Kristina Gligori\'c, Ashna Khetan, Matthias Gerstgrasser, Diyi Yang, and Dan Jurafsky. "Grounding Gaps in Language Model Generations." In \emph{Proceedings of the 2024 Conference of the North American Chapter of the Association for Computational Linguistics: Human Language Technologies}, 6279--6296. 2024.

Silver, David, Aja Huang, Chris J. Maddison, Arthur Guez, Laurent Sifre, George van den Driessche, Julian Schrittwieser, et al. ``Mastering the Game of Go with Deep Neural Networks and Tree Search.'' \emph{Nature} 529 (2016): 484--489.

Skantze, Gabriel, and A. Seza Do\u{g}ru\"oz. "The Open-domain Paradox for Chatbots: Common Ground as the Basis for Human-like Dialogue." arXiv:2303.11708 (2023).

Sperber, Dan, and Deirdre Wilson. ``Irony and the Use-Mention Distinction.'' In \emph{Radical Pragmatics}, edited by Peter Cole. New York: Academic Press, 1981.

Stalnaker, Robert. ``Common Ground.'' \emph{Linguistics and Philosophy} 25 (2002): 701--721.

Su, Miao, Yucan Guo, Zhongni Hou, Long Bai, Zixuan Li, Yufei Zhang, Guojun Yin, Wei Lin, Xiaolong Jin, Jiafeng Guo, and Xueqi Cheng. "Beyond Dialogue Time: Temporal Semantic Memory for Personalized LLM Agents." arXiv:2601.07468, 2026.

Surov, Ilya A., et al. ``Quantum Semantics of Text Perception.'' \emph{Scientific Reports} 11 (2021): 4193.

Tan, Zhen, Jun Yan, I-Hung Hsu, Rujun Han, Zifeng Wang, Long T. Le, Yiwen Song, Yanfei Chen, Hamid Palangi, George Lee, Anand Iyer, Tianlong Chen, Huan Liu, Chen-Yu Lee, and Tomas Pfister. "In Prospect and Retrospect: Reflective Memory Management for Long-term Personalized Dialogue Agents." In \emph{Proceedings of the 63rd Annual Meeting of the Association for Computational Linguistics}, 8416--8439. 2025.

Templeton, Adly, et al. ``Scaling Monosemanticity: Extracting Interpretable Features from Claude 3 Sonnet.'' Transformer Circuits Thread, Anthropic, 2024.

Ueno, Miki. ``Private Etymology: Designing Relational Reuse of Shared Symbols in Long-Term Human--AI Interaction.'' arXiv:2608.08443 (2026).

Vaswani, Ashish, Noam Shazeer, Niki Parmar, Jakob Uszkoreit, Llion Jones, Aidan N. Gomez, \L ukasz Kaiser, and Illia Polosukhin. ``Attention Is All You Need.'' \emph{Advances in Neural Information Processing Systems} 30 (2017): 5998--6008.

Veltman, Frank. ``Defaults in Update Semantics.'' \emph{Journal of Philosophical Logic} 25 (1996): 221--261.

Vermeule, Blakey. \emph{Why Do We Care about Literary Characters?} Baltimore: Johns Hopkins University Press, 2010.

Wang, Zheng, Tyler Solloway, Richard M. Shiffrin, and Jerome R. Busemeyer. ``Context Effects Produced by Question Orders Reveal Quantum Nature of Human Judgments.'' \emph{Proceedings of the National Academy of Sciences} 111 (2014): 9431--9436.

Wilson, Deirdre, and Dan Sperber. ``Explaining Irony.'' In \emph{Meaning and Relevance}. Cambridge: Cambridge University Press, 2012.

Yeshurun, Yaara, Stephen Swanson, Erez Simony, Janice Chen, Christina Lazaridi, Christopher J. Honey, and Uri Hasson. ``Same Story, Different Story: The Neural Representation of Interpretive Frameworks.'' \emph{Psychological Science} 28, no. 3 (2017): 307--319.

Y\i ld\i r\i m, Alper, and \.{I}brahim Y\"uceda\u{g}. ``Language as a Wave Phenomenon: Semantic Phase Locking and Interference in Neural Networks.'' arXiv:2512.01208, 2025.

Yin, Meng. "A quantum-cognitive approach to dynamic meaning construction." \emph{Frontiers in Psychology} 17 (2026): 1664747.

Yu, Wanying, Boyang Ma, Zhibo Eric Sun, Minghui Xu, and Yue Zhang. "Bad company corrupts good morals: Understanding and Measuring Narrative-Induced Moral Reasoning Degradation in LLMs." arXiv:2606.28981v1 {[}cs.CY{]}, June 27, 2026.
\end{hangparas}

\end{document}